\documentclass[10pt, journal]{article}
\pdfoutput=1
\usepackage{blindtext}
\usepackage{graphicx} 
\graphicspath{{/home/mike/Dropbox/Latex/MK_images/Graphics1/}}
\DeclareGraphicsExtensions{.png, .jpg, .pdf}
\usepackage{amsmath} 
\usepackage{amsfonts}
\usepackage{bbm}
\usepackage{subfig}
\usepackage{float}
\usepackage[utf8]{inputenc}
\usepackage{graphicx}
\usepackage{lipsum}
\usepackage[margin={2cm,2cm}]{geometry}
\usepackage{amsthm}
\theoremstyle{plain}

\usepackage{url}
\usepackage{multicol}
\usepackage{vwcol}
\usepackage{wrapfig}
\usepackage{authblk}

\begin{document}

	
%
\title{The Monge-Kantorovich Optimal Transport Distance for Image Comparison}
\date{}
\author[]{Michael Miller}
\author[]{Jan Van lent}
\affil[]{\small\textit{Research Group in Applied Mathematics and its Applications \&} \\ \small\textit{Centre for Machine Vision} \\ \small\textit{University of the West of England, Bristol, BS16 1QY}}
\affil[]{\textit{\{Michael1.Snow\}@outlook.com}, \textit{\{Jan.Vanlent\}@uwe.ac.uk}}

\maketitle

\begin{multicols}{2}

\begin{abstract}
\textit{This paper focuses on the Monge-Kantorovich formulation of the optimal transport problem and the associated $L^2$ Wasserstein distance. We use the $L^2$ Wasserstein distance in the Nearest Neighbour (NN) machine learning architecture to demonstrate the potential power of the optimal transport distance for image comparison. We compare the Wasserstein distance to other established distances - including the partial differential equation (PDE) formulation of the optimal transport problem - and demonstrate that on the well known MNIST optical character recognition dataset, it achieves excellent results.}
\end{abstract}


\section{Introduction}

The topic of optimal transport is an important research topic in a number of areas including economics, fluid mechanics and computational mathematics. However, optimal transport seems to be a little known technique for the direct comparison of images outside of image registration. In the main, this is due to the time costs of solving the linear programming problem that's set out in Section IV. It has been widely used for one-dimensional problems - such as histogram comparison - and is commonly known as the `Earth Mover's Distance' (EMD) where the $L^2$ distance is used in formulating the optimal transport problem. In this paper we would like to demonstrate that the $L^2$ Wasserstein distance - that arises from solving the optimal transport problem - is a natural way to compare images and indeed can be a more natural choice of metric than other established distances. 

To introduce and motivate the topic of optimal transport, we shall present it from an economics point of view - and show its link to image comparison. Imagine we have a set of factories at locations in a city. Each factory holds a quantity of goods. We also have a set of retailers at different locations in a city whose total demand of goods is equal to the total supply of goods. If we assume that to transport the goods from a factory to a retailer has a cost attached to it (for instance fuel, employment costs, et cetera), then the economic problem becomes `How do we transport the goods from the factories to the retailers whilst minimising the transportation cost?'. 

\begin{figure}[H]
	\begin{center}
		\includegraphics[width=0.45\textwidth, height = 0.28\textheight]{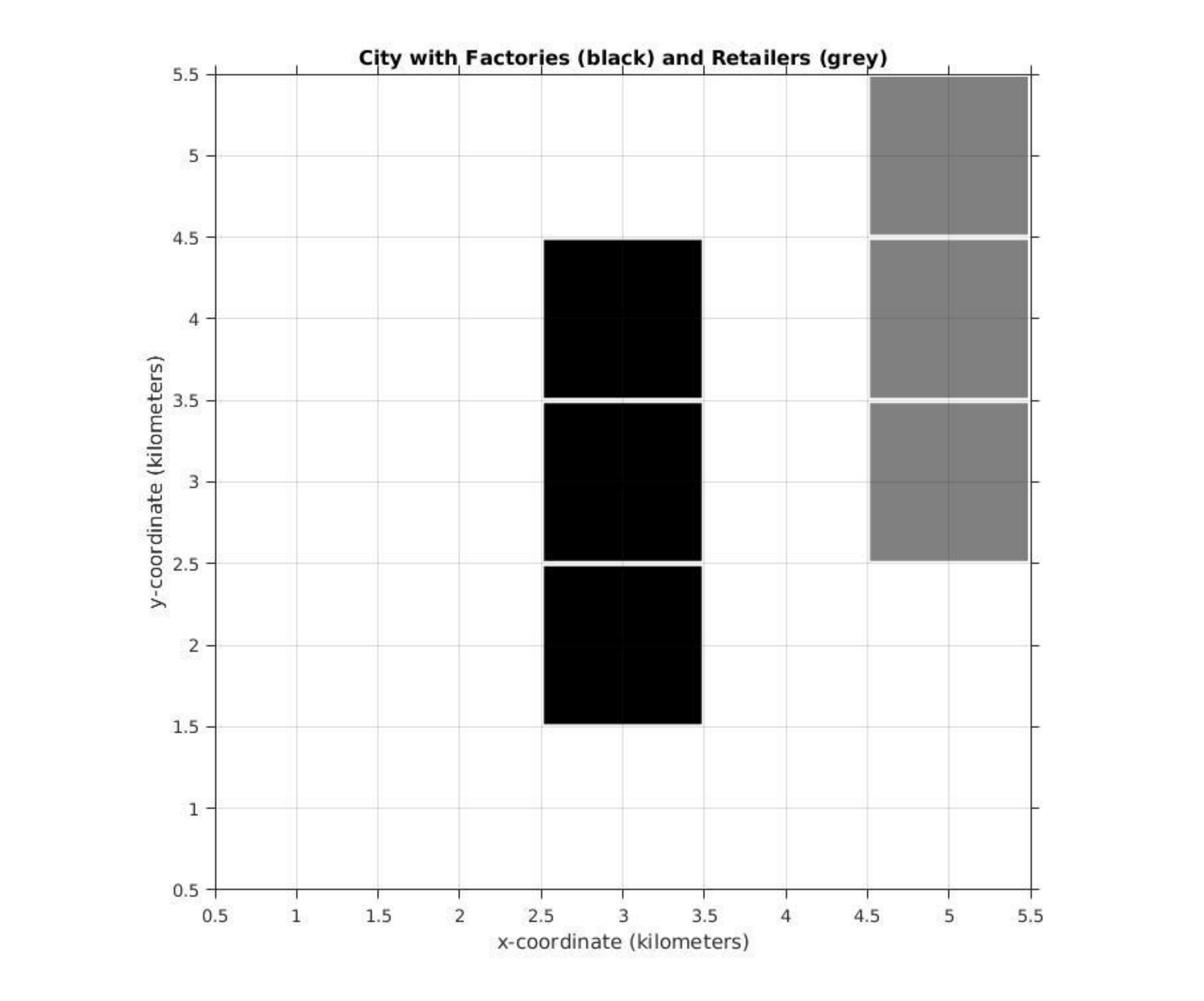}
		\caption{Factory to Retailer Example}
		\label{city}
	\end{center}
\end{figure}

As a simple example, we shall consider 3 bakers (in black) each holding a single loaf of bread and 3 caf\'es (in grey) each with a supply of one loaf of bread. The geometry is shown in Figure \ref{city}, with each baker and caf\'e at the centred in the middle of the square. We shall assume that the cost of transporting goods is proportional the square of the distance travelled. 

Clearly in this case, the optimal way to transport the goods is a translation. The cost of this transportation will be squared distance of the translation for each of the three loaves of bread multiplied by the number of goods - which is just 1 in this example. For example, from the baker centred at $(x,y) = (3,2)$, we transport to the retailer centred at $(x,y) = (5,3)$, and the distance is the hypotenuse of a right triangle $\sqrt{5}$; hence the squared distance is $5$. Transportation from the other two bakers is analogous, hence the total transportation cost is 15. A formal presentation of the problem is given in Section IV.

This transportation cost can be adapted to a metric for images and is known as the $L^2$ Wasserstein distance in the case of the cost being proportional to the distance squared. Instead of considering an economics problems, we can think of two greyscale images. We consider one to be source domain, and one to be a target domain, where pixel intensities are densities. This is analogous to the economics problem where one domain is the supply and one domain is the supply, when densities relate to supply and demand. Typically, the sum of the pixels of two greyscale images will not be equal (i.e. supply not being equal to demand) and so we have to adapt the problem. In this paper, we will normalise the images such that they have equal pixel sums. The image comparison problem can then be thought of as `What is the most efficient way to morph one image onto another?'.  In many cases this will be a more natural way of comparing images than classical techniques. In this simple example, if we had one image containing the bakers and another image containing the caf\'es, if we used the Euclidean distance to compare the images for instance, this would produce poor results as the two regions of interest are disjoint. This would not give a good metric for image similarity when we consider that the images look visually similar. 

\section{Related Work}

There is a limited amount of literature on using the optimal transport distance to compare images in two dimensions or show how well it can perform in image comparison tasks. There is more literature available on the more well-known EMD; see for example Ling \cite{HL01} and Rubner \cite{YR01}. This is the $L^1$ Monge-Kantorovich distance for distributions of equal weight; commonly used to compare histograms. The first use of the Wasserstein distance in two dimensions is by Werman et al. \cite{MW01, MW02}. In this case Werman used a discrete `match distance' which is the Monge-Kantorovich distance and demonstrated a case where images of equal pixel sum could be compared. Werman noted that the distance has been shown to have many theoretical advantages but a significant drawback is the computational cost. For instance naively comparing two $n \times n$ images would be of complexity $O(N^3)$ according to Werman, where $N$ is the number of pixels. In the paper by Kaijser \cite{TK01}, Kaijser managed to reduce the complexity of the discrete Wasserstein algorithm from $O(N^3)$ to roughly $O(N^2)$, where $N$ denotes the number of pixels, by considering \textit{admissible arcs} in which to transport from one image to another. Kaijser's paper focuses on constructing a more efficient algorithm rather than the relative benefits of using the Wasserstein distance, but he does illustrate the distance being used to compare images. In a similar vein, Alexopoulos \cite{CA01} claims an improvement upon Kaijser's algorithm and also presents some example image comparisons. There have been some notable contributions to the field of optimal transport and image processing by Delon et al. \cite{JD01, JD02}. Delon has implemented some novel approaches to transporting the colour maps from one image onto another. Delon has also worked in developing fast solvers to the transport on different domains, for instance solving a transport problem on a cricle by cutting the circle and computing the distance on the real line. This can be applied to things such as circular histograms, for instance circular colour spaces. One dimensional problems are preferable if computational complexity is a consideration. The northwest corner method can give a unique optimal solution with a strictly convex cost function for instance. Snow and Van lent \cite{MM01} have implemented an algorithm that compares images using Monge's PDE formulation of the optimal transport distance for images and found it to have excellent results when using it in the 1-NN algorithm. Indeed, this is the motivation for this paper. We would expect the Monge-Kantorovich distance to produce similar results to Snow and Van lent's paper on the same dataset and we would like to draw attention to the advantages and disadvantages of each formulation. 

\subsection*{Contributions and Outline}

The main contributions of this paper are as follows. The paper sets out the Monge-Kantorovich formulation of the optimal transport problem and an approach to solve this for image comparison. We demonstrate the usefulness of this optimal transport distance by utilising it in a machine learning context to classify images. In this case, we shall use NN, which is detailed in Section V, equipped with the optimal transport distance. We shall compare this both to a standard Euclidean distance and the excellent Tangent Space Distance (TSD). We also compare it to a PDE formulation of the optimal transport problem to draw attention to the advantages and disadvantages of using this approach comparing to the PDE formulation. In this paper we are not interested in the computational complexity of the Monge-Kantorovich problem.

We set this out as follows. In Section III and IV we outline the optimal transport problem and the Monge-Kantorovich formulation of the optimal transport problem; including an outline of the algorithm to solve an image comparison problem. In Section V we outline the NN algorithm and the Euclidean, Tangent Space and optimal transport distances we will be using in the NN classifier. In Section VI we give our methodology for comparing images and illustrate our approach using the well-known MNIST dataset. In Section VII we present our results and finally in Section VIII, we provide a summary of the work.

\section{Monge's Optimal Transport Problem}

Gaspard Monge was the founding father of the study of optimal transport. Monge's problem is as follows: consider a finite pile of rubble (mass) and a hole of the exact same volume. Assume that to transport the dirt to the hole requires some measure of effort, then Monge's problem is to find an optimal mapping to transport the mass into the hole such that the effort is minimised \cite{CV01}. 

\begin{figure}[H]
	\begin{center}
		\includegraphics[width=0.45\textwidth, height = 0.15\textheight]{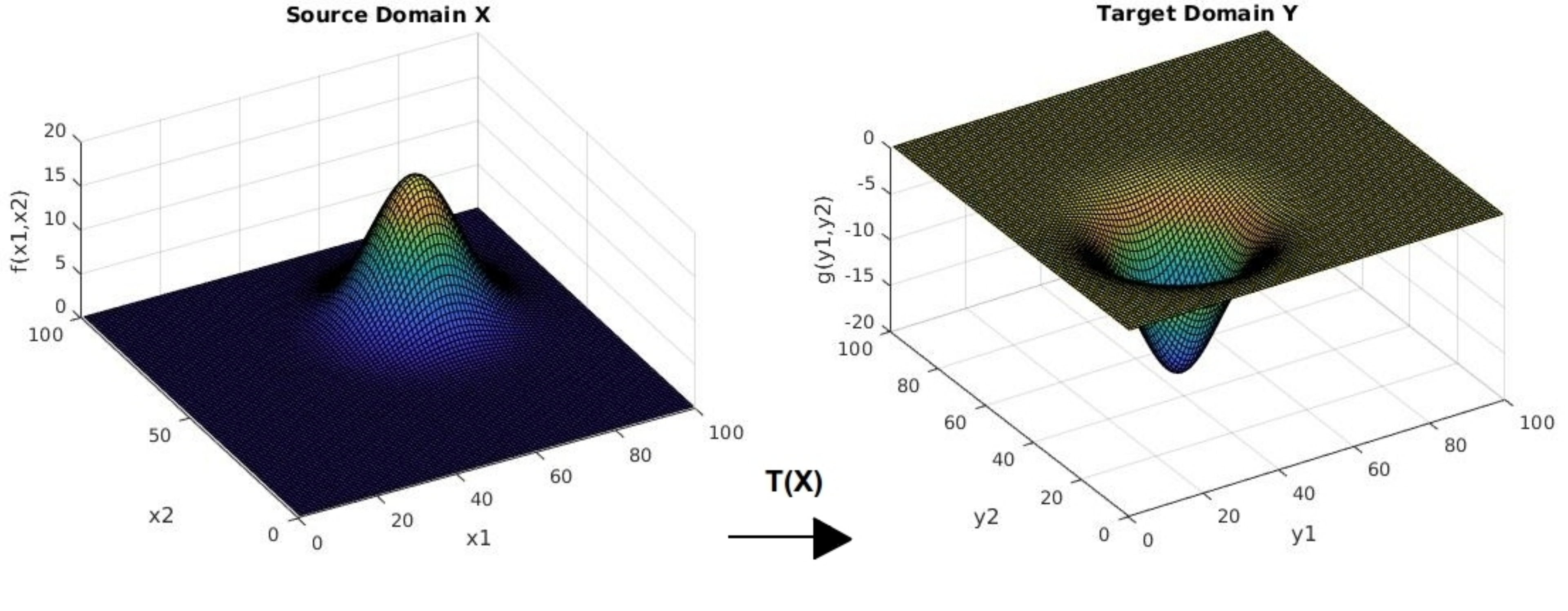}
		\caption{Optimal Transport Motivation}
		\label{dirt}
	\end{center}
\end{figure}

The transport problem can be described mathematically as follows. Consider two measure spaces $(X,\mu)$ and $(Y,\nu)$, where $X$ is the source domain and $Y$ is the target domain. The measures $\mu$ and $\nu$ model the pile of dirt and the hole respectively. By normalising such that the volume of the pile of dirt and hole integrate to 1, $\mu$ and $\nu$ can be considered probability measures (more generally Radon measures on a Polish Space). We denote $\mu(A)$ and $\nu(B)$ to represent the mass of measurable subsets $A \subset X$ and $B \subset Y$. We consider that to transport the mass from some $x \in X$ to some $y \in Y$ requires effort. We prescribe a cost function to measure the effort of transporting mass from $X$ to $Y$. It is natural to consider this cost function as a non-negative distance metric as we are describing effort proportional to distance. So we shall define $c: X \times Y \rightarrow [0,+\infty]$, where $c$ is measurable. 
 
Monge's problem is to find a bijective mapping $T: X \rightarrow Y$ that rearranges - or transports - one density onto the other. For the mapping $T$ to be valid, we must have that mass is preserved by the mapping, hence

\begin{align}
\nu(A) = \mu(T^{-1}(A)), \qquad \qquad \forall A \subset X.
\end{align}

This is sometimes denoted $\nu = T\#\mu$,  the \textit{push-forward} of $\mu$ by $T$ onto $\nu$. Amongst all such valid maps, we wish to minimise the effort with respect to the cost function and so Monge's problem can be stated as minimising the following:

\begin{align}
I[T] = \int_X c(x,T(x))d\mu(x), \qquad \text{s.t.} \qquad \nu = T\#\mu.
\end{align}

\begin{figure}[H]
	\begin{center}
		\includegraphics[width=0.47\textwidth, height = 0.15\textheight]{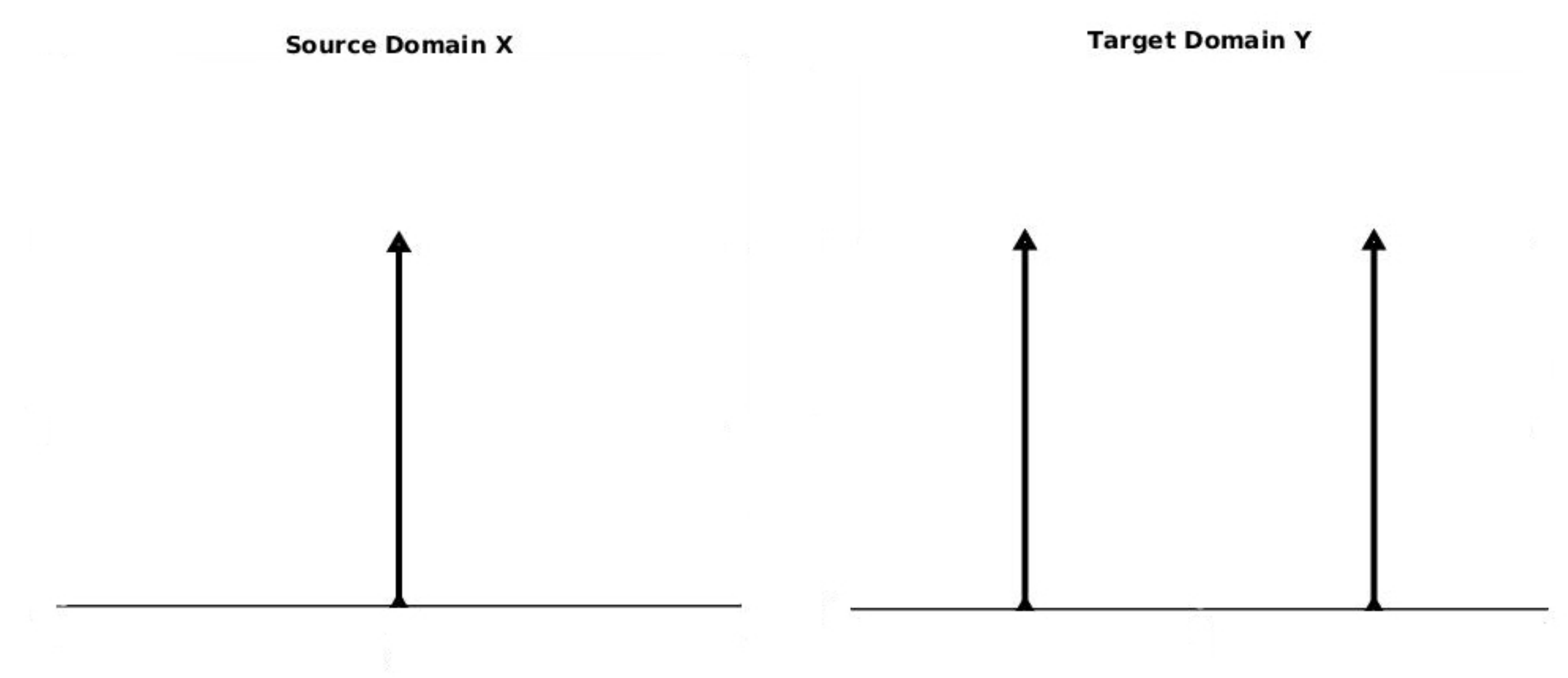}
		\caption{Example where a mapping $T(x)$ cannot exist.}
		\label{Dirac}
	\end{center}
\end{figure}

There are difficult constraints on the mapping $T$: Monge's problem is highly non-linear and can potentially be ill-posed. There may in fact be instances where there does not exist a mapping at all; for example a simple case is to consider when $\mu$ is a Dirac measure $\mu = \delta_0$ but $\nu$ is not. Trivially, no mapping can exist as mass cannot be split. This is illustrated in Figure \ref{Dirac}. 

\section{The Monge-Kantorovich Optimal Transport problem}

Kantorovich proposed a relaxation of Monge's formulation which quite remarkably led to a linear problem. Instead of finding a mapping, $T$, between the two densities, Kantorovich constructed a transport plan $\pi(x,y)$, which tells us the amount of mass transported from $x \in X$ to $y \in Y$. In this case the mass can split. More formally, we consider two Polish spaces, equipped with measures $(X,\mu)$ and $(Y,\nu)$, where $X$ is the source domain and $Y$ is the target domain. We consider a measurable cost function $c: X \times Y \mapsto [0,+\infty]$ which prescribes a cost to transport the mass from $X$ to $Y$. To transport the mass we consider a transport plan $\pi(x,y)\geq0$, where $\mu$ is a measure, on the product space $X \times Y$, where mass can split. For the plan to be admissible, the mass must be preserved under the plan. This gives the constraints

\begin{align}
\int_Y d&\pi(x,y) = d\mu(x), \qquad \int_X d\pi(x,y) = d\nu(y) \nonumber\\
& \pi(x,y) \geq 0. 
\end{align}

Over all such admissible plans, the optimal transport is now the following primal linear programming problem:

\begin{align}
I[\pi] = \inf \left\{\int_{X \times Y} c(x,y)d\pi(x,y): \, (3) \,  \text{is true}\right\}. 
\end{align}

This has the associated dual formulation. If we consider the test functions $u: X \rightarrow \mathbb{R}$ and $v: Y \rightarrow \mathbb{R}$, the dual is given by 

\begin{align}
I = \sup \left\{\int_X u(x)d\mu(x) + \int_Y v(y)d\nu(y)\right\},
\end{align}

such that

\begin{align}
u(x) + v(y) \leq c(x,y). 
\end{align}

Typically, we will assume that the densities are absolutely continuous with respect to Lebesgue measure, i.e. smooth \cite{CV01}. In this case we have density functions such that

\begin{align}
d\mu = f(x)dx, \quad \text{and} \quad d\nu= g(y)dy.
\end{align}

Our primal problem becomes 

\begin{align}
I[\mu] = \min \int_{X \times Y} c(x,y)d\pi(x,y)dxdy,
\end{align}

subject to the constraints 

\begin{align}
\int_Y d\pi(x,y)dy = f(x), \qquad \int_X d\pi(x,y)dx = g(y), \qquad 
\end{align}

with 

\begin{align}
d\pi(x,y) \geq 0.
\end{align}

The corresponding dual problem becomes a maximisation of the objective function

\begin{align}
\int_X u(x)f(x)dx + \int_Y v(y)g(y)dy,
\end{align}

such that 

\begin{align}
u(x) + v(y) \leq c(x,y). 
\end{align}

The common approach to finding a solution to this continuous problem is to discretise the primal formulation by considering the densities to be the sum of Dirac measures. If we have  such that 

\begin{align}
f(x) &= \sum\limits_{i=1}^m \delta(x_i)f(x_i) 
\end{align}
and
\begin{align}
g(y) &= \sum\limits_{j=1}^n \delta(y_j)g(y_j),
\end{align}

where $\delta_k = \delta(x-x_k)$ is the Dirac Delta function. This discrete problem, known as the discrete Monge-Kantorovich optimal transport problem, is now a finite-dimensional linear programming problem. For the problem to be valid, we require the sum of the masses to be equal:

\begin{align}
\sum_{i=1}^{m} f(x_i) = \sum_{j=1}^{n} g(y_j)
\end{align}

We also construct a cost vector $c_{ij}$. The vector $c_{ij}$ measures a chosen distance between the points $\mu_i^+$ and $\mu_j^-$ at the points $x_i$ and $y_j$. This is to measure the cost of transporting mass from $X$ to $Y$. We shall choose our cost function to be the squared distance between points $c_{ij} = \lvert x_i - y_j \rvert^2$, where $x_i$ and $y_j$ in this case denote the point positions. This is an approximation to the $L^2$ Wasserstein distance and so we can compare to the PDE approach set out by Miller and Van lent \cite{MM01}. The problem then becomes a minimisation: we wish to find some non-negative transport plan $\pi_{ij} \geq 0$ so that we minimise

\begin{align}
\sum_{i=1}^n \sum_{j=1}^n c_{ij} \pi_{ij}, 	
\end{align}

subject to the constraints

\begin{align}
\sum_{j=1}^n \pi(x_i,y_j) = f(x_i) \qquad \text{and} \qquad \sum_{i=1}^m \pi(x_i,y_j) = g(y_j).  
\end{align}

This can be written as a standard primal linear programming (LP) problem \cite{LE01}. If $x = \left(\mu_{11}, \mu_{12}, ... \, , \mu_{nn}\right)^T$, then the LP formulation is

\begin{align*}
&\textbf{Minimise} \,\, c \cdot x, \,\, \text{subject to the constraints} \\
&A\pi = b, \,\, \pi \geq 0, 
\end{align*}

where $b = \left(f(x_1), ... \, , f(x_m), g(y_1), ... \, , g(y_n)\right)^T$  and $A$ is the constraint matrix of the form 

\begin{align}
A = \begin{bmatrix}  \mathbbm{1} & 0 & \dots & 0 \\
0 & \mathbbm{1} & \dots & 0 \\
\vdots & \vdots & \dots & \vdots \\
0 & 0 & \dots & \mathbbm{1} \\
e_1 & e_1 & \dots & e_1 \\
e_2 & e_2 & \dots & e_2 \\
\vdots & \vdots & \dots & \vdots \\
e_n & e_n & \dots & e_n \\
\end{bmatrix}.
\end{align}

Here, $\mathbbm{1} \in \mathbb{R}^m$ is a row vector of ones and $e_i \in \mathbb{R}^n$ is a row vector which contains a $1$ at the $i$th position, else contains a zero. The can be solved using standard LP solvers, but the typical cost of solving the transport problem naively is at least $O(N^3)$ where $N$ is the number of pixels. 

\section{Nearest Neighbour Classification}

One of the simplest machine learning algorithm is the nearest neighbour algorithm. The NN algorithm is an example of non-parametric instance-based learning \cite{NA01}; given a set of labelled training data and a set of labelled test data, each instance of test data is compared to every instance of training data subject to some similarity measure. The test point is classified as the class of the nearest neighbour to the test point.

More formally, we consider general data points to be in a metric space $(X,d)$, equipped with a distance $d$, where each data point has an associated class. There is no limit to the amount of classes; the algorithm works equally well for more classes. We then choose some $k \in \mathbb{N}^+$ to define the number of nearest neighbours that influence the classification. In the simplest case $k = 1$, which is our choice in this paper, the classification rule is that the training vector closest to the test vector under the distance $d$ determines the class of the test vector. 

\subsection{Distances}

In this experiment, we shall use the NN algorithm equipped with the Monge-Kantorovich distance and compare it to three other distances to assess its effectiveness. The first we have chosen is the well-known Euclidean distance. Given two $n$-dimensional vectors to compare $\mathbf{x}$ and $\mathbf{y}$, the Euclidean distance is defined as

\begin{align*}
d(\mathbf{x}, \mathbf{y}) = \sqrt{(x_1 - y_1)^2 + (x_2 - y_2)^2 + \cdot \cdot + (x_n - y_n)^2}.
\end{align*}

The second distance we shall compare against is the Tangent Space Distance, which produces excellent results on optical character recognition datasets with the NN framework. Simard et al. \cite{PS01} developed a distance measure that is invariant to small transformations of images, something that other distances such as the Euclidean distance are very sensitive to. Simard et al. approximated the small transformations by creating a tangent space to the image by adding the image $\mathcal{I}$ to a linear combination of the tangent vectors $t_l(x)$, for $l = 1,.\, . \, , L$, where $L$ is the number of different transformation parameters. The tangent vectors are the partial derivatives of the transformation with respect to the parameter and span the tangent space. So the tangent space is defined as

\begin{align} 
M_x = \{ \mathcal{I} + \sum\limits_{l = 1}^L t_l(x)\cdot \alpha : \alpha \in \mathbb{R}^L \},
\end{align}

where $\alpha$ is the transformation parameter, for example the scale or translation. The one-sided TSD is they defined as 

\begin{align}
TSD\left(\mathcal{I}_1, \mathcal{I}_2\right) = \min\limits_{\alpha} \{ \lVert \mathcal{I}_1 + \sum\limits_{l = 1}^L t_l(x)\cdot \alpha - \mathcal{I}_2 \rVert \}.
\end{align}

In our experiment we have used Keyser's implementation of the TSD \cite{DK01}. 

Finally we shall compare to the PDE formualation of the optimal transport problem. In the case of choosing the $L^2$ distance as the cost function - as we have with the Monge-Kantorovich distance - work by Brenier et al. \cite{YB01} has shown that under certain conditions, the optimal transport mapping is given by the gradient of a displacement potential function $\phi: \mathbb{R} \mapsto \mathbb{R}$ such that $\phi = \frac{1}{2}\lvert x \rvert^2 - u$. This leads to solving a non-linear PDE

\begin{align} 
\det(I-\nabla^2 u)g(x-\nabla u) = f(x),
\end{align}

subject to boundary conditions ensuring $T(X) = Y$ when mapping a square to itself, see for example \cite{CV01}. Snow and Van lent \cite{MM01} presented a numerical solution to this formulation in the case where the density functions are images. This distance is an approximation to the $L^2$ Wasserstein distance, so we would expect the results to be similar given both are approximations to the distance. 

\section{Nearest Neighbour Classification on the MNIST dataset}

\begin{figure}[H]
	\begin{center}
		\includegraphics[width=0.45\textwidth]{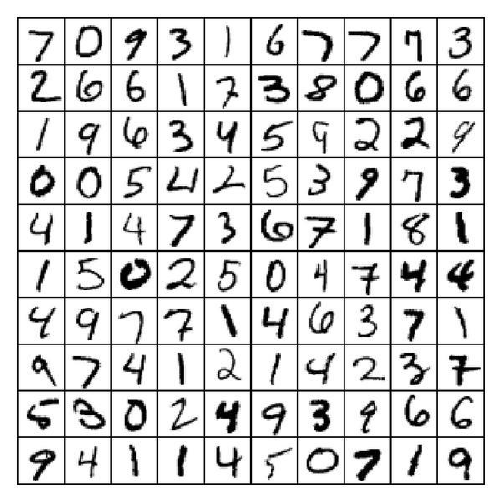}
		\caption{Examples of MNIST Digits}
		\label{mnist1}
	\end{center}
\end{figure}

The MNIST digit dataset is a well-known dataset in the field of digit recognition constructed by LeCun et al. \cite{YL01} from the NIST dataset. According to LeCun et al. the original binary images were size normalised to fit into a 20 $\times$ 20 pixel box while preserving their aspect ratio. The MNIST dataset are a grey level representation of the images, resulting from the anti-aliasing algorithm used in their normalisation. The images are centred into a $28 \times 28$ image by computing the centre of mass of the pixels and translating the images such that this point coincides with the centre of the $28 \times 28$ image. The dataset contains a training set with 60,000 images and a test set of 10,000 images. This is a convenient dataset to test our algorithm on as it requires minimal preprocessing and the images are centred so we can focus on the distance metric's discriminatory power over translation and scaling invariance. Figure \ref{mnist1} examples some of the digits. Although this is a nice dataset in that it requires minimal preprocessing, it is worth noting that this is not the only choice one can make in the preprocessing steps. For example, if the images are not centred by their mass and just drawn within some bounding box, the results of a nearest neighbour classification may significantly differ depending on the properties of the distance metric chosen.

This is a strong test of how well a distance metric can perform given the natural deviations in how people write digits. In particular, some of the digits can have a very similar pixel distribution. 

In our experiment it is worth noting that we used the exact same dataset as Snow and Van lent \cite{MM01} such as to directly compare the results. The dataset is constructed as follows. We have taken a subset of the MNIST dataset to perform our NN test. We first randomised the original dataset and then created a training set of 10,000 images which contained 1,000 different examples of each of the digits. This was further partitioned into 20 disjoint training sets containing 21 samples of each digit. A test set was randomly constructed from the MNIST test set which contained 200 images, with 20 of each digit. Both the training set and the test sets were normalised such that that each the total pixel sum of each image was equal to 1, as required in the Monge-Kantorovich formulation we set out.

We were interested to see how each metric performed as the training set increased in size, starting from a very small training set (see for example \cite{AS01}). We split our experiment into 21 different tests; for the initial test we performed the NN algorithm with a training set of just 1 of each digit. We repeated this test on each of the four distances to be tested, each time increasing the number of each digit in the training set by 1. E.g. for the final test, we had a training set containing 21 of the digit `one', 21 of the digit `two' and so on. This experiment was then tested on each of the 20 different constructed training sets to take into account the variability of the data sets. 

In this experiment, we shall impose the same conditions on the dataset for each of the distance measures. We define $k=1$ for the nearest neighbour classification. In the case of the PDE formulation of the optimal transport problem, to ensure that the numerical solution method works for every image with the same parameters, we added a constant of $1$ to each image before normalisation to ensure every image is both non-zero and smooth. In this case we can use Newton's method without a damping parameter mentioned in the work by Miller and Van lent \cite{MM01}.

\section{Results}

In this section we shall present our main results. The main result we captured was the accuracy of each tested distance averaged over all 20 of the disjoint training sets. This main result is as we expected. The Kantorovich distance approximately has the same results as the PDE formulation and performs better than both the Euclidean distance and the TSD. There is a slight discrepancy between the results of the PDE formulation and the linear programming formulation of the optimal transport problem. We suggest that this is down to the approximations each algorithm has made; for instance the PDE formulation uses cubic interpolation. Generally, the results are very similar.

\begin{figure}[H]
	\begin{center}
		\includegraphics[width=0.50\textwidth, height = 0.32\textheight]{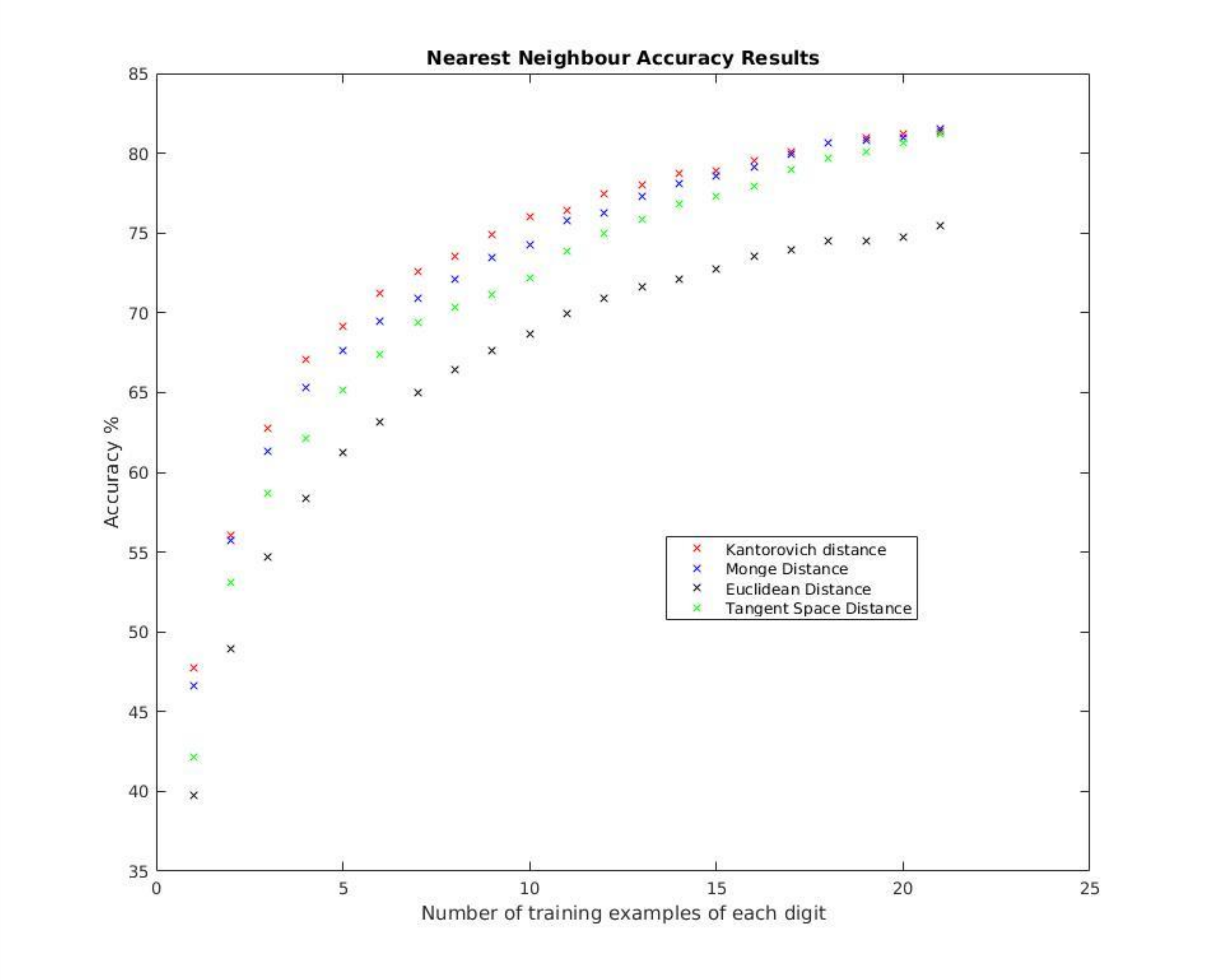}
		\caption{- An average of the results over all 20 training sets.}
		\label{Euclid_results}
	\end{center}
\end{figure}

\begin{figure}[H]
	\begin{center}
		\includegraphics[width=0.5\textwidth, height = 0.32\textheight]{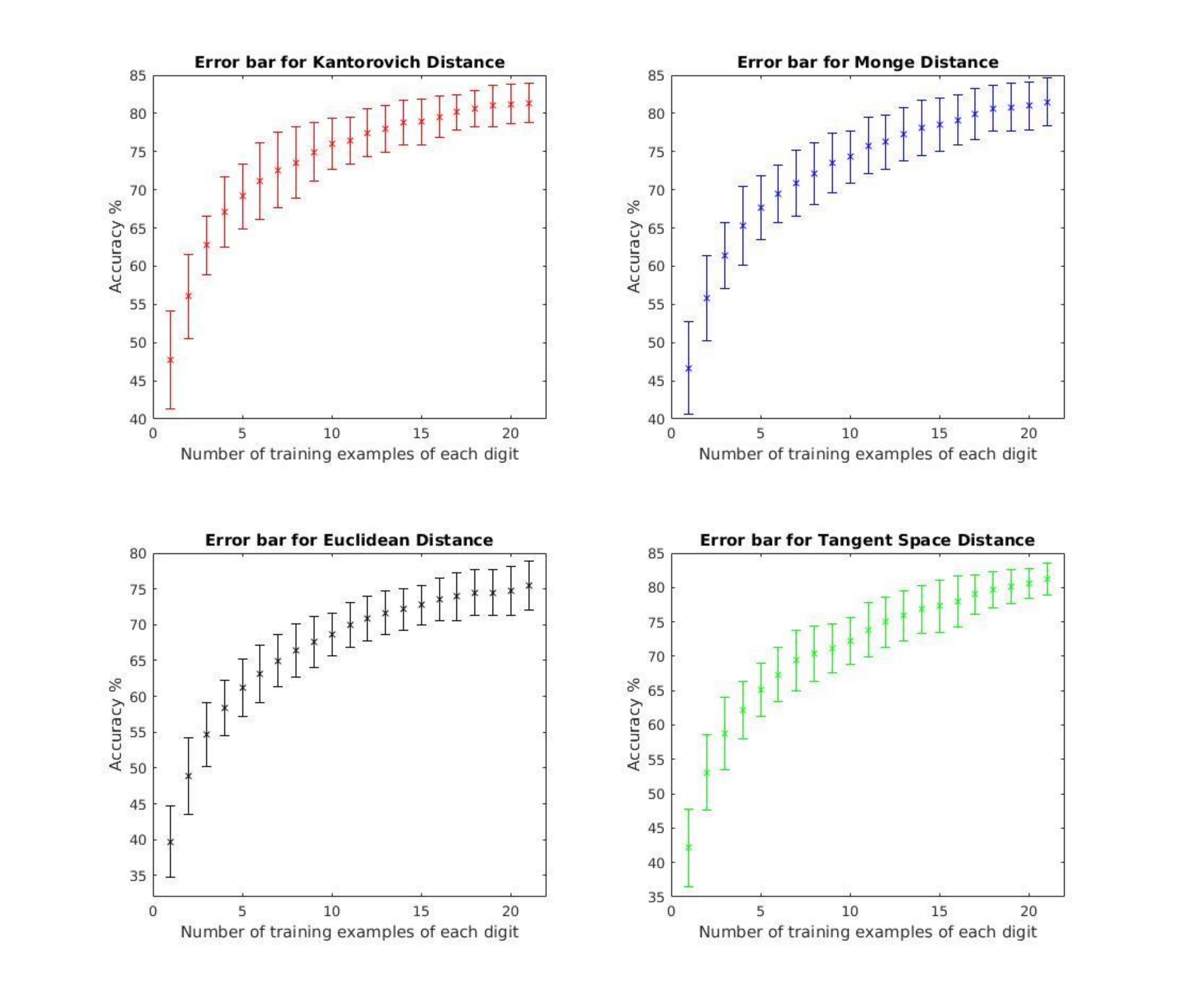}
		\caption{- An error bar of one standard deviation for each of the distances over the 20 data sets.}
		\label{Pearson_results}
	\end{center}
\end{figure}

The overall results are presented in Figures \ref{Euclid_results}. These show the average accuracy over all 20 tests conducted. We have also included the error bars for each of the distances in Figure \ref{Pearson_results}; the error bars present one standard devation from the mean. We can see that the variability is generally very similar in each of the distances tested.

We also include a sample of the results in Table \ref{Results Table} to complement the graphics illustrating the results. The results show that both formulations of the optimal transport distance outperform the other distances tested. The PDE formulation shows a slight improvement on the Monge-Kantorovich formulation for a larger training set. The results amplify what has already been concluded by Snow and Van lent in that this demonstrates that optimal transport is a natural way to compare images in some applications.

\begin{table}[H]
	\caption{Table of Sampled Results}
\begin{tabular}{ |p{2cm}|p{0.75cm}|p{0.75cm}|p{0.75cm}|p{0.75cm}|p{0.75cm}|}
	\hline
	\multicolumn{6}{|c|}{Number of Training Digits with Accuracy (\%) 1 d.p.} \\
	\hline
	Distance & 1 & 5 & 10 & 15 & 21\\
	\hline
	Euclidean  & 37.5 & 58.8 & 66.3 & 71.6 & 75.5\\ 
	Tangent     & 42.1 & 65.1 & 72.2 & 77.3 & 80.6\\
	Kantorovich & 47.7 & 69.2 & 76 & 78.9 & 81.4 \\ 
	Monge & 45.2 & 68.2 & 75.1 & 79.6 & 82.6\\ 
	\hline
\end{tabular}
\\[9pt]
\caption*{A sample of the results to relate to Figure 12.}
\label{Results Table}
\end{table}	

\subsection{Comparison of optimal transport formulations}

Whilst both formulations of the optimal transport problem yield similar results, it is worth taking about their relative merits. The Monge-Kantorovich formulation is both very easy to understand and formulate. As shown in Section III, this involves simply setting up the constraint matrix to formulate the problem as a standard linear programming problem. To solve the problem, a standard LP solver can be used, such as a simplex method or interior point method. The simplicity is a big benefit if the user is not concerned about the computational time and simply wishes to know some metric between images. Complexity is a significant drawback, though. For example, trying to compare high resolution images in a machine learning context would not be practical; even with recent advances in the efficiency of the algorithm. The PDE formulation certainly benefits in computation time, but the implementation is significantly more challenging. There are many considerations to take into account, such as: how to interpolate the images; normalisation; acceptable error. In either case, we have shown that in some cases, using the optimal transport distance would certainly be beneficial in some areas of image processing and machine learning. 

\section{Summary and Future Work}

In this paper we have shown an implementation of the Monge-Kantorovich optimal transport problem with which to compare images. As expected, the algorithm performs well as the PDE formulation of the optimal transport problem and against other leading distance metrics. Although there are some drawbacks in computation time when using the Monge-Kantorovich distance, it is a relatively easy formulation and implementation in comparison to PDE formulation - so in many cases this may be an appropriate choice. Future work will be to look at the accuracy of the discretisation of the Monge-Kantorovich problem and to investigate the structure of the formulation to increase the speed.

\section{Acknowledgements}

We would like to thank the University of the West of England's Centre for Machine Vision research group, where this research was conducted.   

\bibliography{/home/mike/Dropbox/Latex/MK_images/mybib}

\bibliographystyle{ieeetr}

\end{multicols}
\end{document}